\documentclass[fleqn,10pt]{wlscirep}
\usepackage[utf8]{inputenc}
\usepackage[T1]{fontenc}
\usepackage{lineno}
\usepackage{multirow}
\usepackage{threeparttable}
\usepackage{tabularx}
\usepackage{float}
\usepackage{hyperref}
\usepackage{xurl}

\ifdefined\showrevision
\usepackage{xcolor}
\usepackage[normalem]{ulem}
\usepackage{marginnote}
\usepackage{xspace}

\newcommand{\revref}[2]{%
\marginnote{$R_{#1}C_{#2}$}
}
\DeclareRobustCommand{\robustrevref}[2]{%
\revref{#1}{#2}
}

\newcommand{\revmod}[1]{%
{\color{blue}#1}\xspace
}

\newcommand{\revnew}[1]{%
{\color{orange}#1}\xspace
}

\newcommand{\revdel}[1]{%
{\color{darkgray}\sout{#1}}\xspace
}

\else
\usepackage{xspace}

\newcommand{\marginnote}[1]{\ignorespaces}
\newcommand{\revref}[2]{\ignorespaces}
\newcommand{\robustrevref}[2]{\ignorespaces}
\newcommand{\revmod}[1]{#1\xspace}
\newcommand{\revnew}[1]{#1\xspace}
\newcommand{\revdel}[1]{\ignorespaces}

\fi
\newcommand{\figref}[1]{\figurename~\ref{#1}}
\newcommand{\tabref}[1]{\tablename~\ref{#1}}

\title{ROBUST-MIPS: A Combined Skeletal Pose and Instance Segmentation Dataset for Laparoscopic Surgical Instruments}

\author[1]{Zhe Han}
\author[1,*]{Charlie Budd}
\author[1]{Gongyu Zhang}
\author[1]{Huanyu Tian}
\author[1]{Christos Bergeles}
\author[1]{Tom Vercauteren}
\affil[1]{King's College London, School of Biomedical Engineering \& Imaging Sciences, London, SE1 7EU, UK}

\affil[*]{corresponding author: charles.budd@kcl.ac.uk }

\begin{abstract} % 170 words max
Localisation of surgical tools constitutes a foundational building block for computer-assisted interventional technologies.
Works in this field typically focus on training deep learning models to perform segmentation tasks. 
Performance of learning-based approaches is limited by the availability of diverse annotated data.
We argue that skeletal pose annotations are a more efficient annotation approach for surgical tools, striking a balance between richness of semantic information and ease of annotation, thus allowing for accelerated growth of available annotated data.
To encourage adoption of this annotation style, we present, ROBUST-MIPS, a combined tool pose and tool instance segmentation dataset derived from the existing ROBUST-MIS dataset.
Our enriched dataset facilitates the joint study of these two annotation styles and allow head-to-head comparison on various downstream tasks.
To demonstrate the adequacy of pose annotations for surgical tool localisation, we set up a simple benchmark using popular pose estimation methods and observe high-quality results.
To ease adoption, together with the dataset, we release our benchmark models and custom tool pose annotation software.
\end{abstract}
\begin{document}

\flushbottom
\maketitle

\thispagestyle{empty}

\section*{Background \& summary} % 700 words max
The localisation of surgical tools in intraoperative endoscopic video is a key capability in computer-assisted intervention (CAI).
It holds the potential to facilitate novel CAI features such as safety analysis~\cite{rios2023cholec80} and automated endoscope control~\cite{gruijthuijsen2022robotic}, whilst also demonstrating a level of surgical scene understanding which could build into more complex technologies.
While localisation can take many forms, the majority of works in this field focus on semantic segmentation, whereby a class label is predicted for every pixel in the image~\cite{ronneberger2015u,8206462}.
Additionally, some works have incorporated instance segmentation~\cite{alabi2025cholecinstanceseg}, a technique that extends semantic segmentation by distinguishing between individual instances of the same object class.
Annotations for semantic segmentation require the creation of complex polygons or curves that follow the contours of each semantic object, be it a tool or a tool-part\cite{twinanda2016endonet,hong2020cholecseg8k,ross2021comparative}.
While these annotations provide detailed semantic information, they require significant time to create.
In general-purpose computer vision domains, bounding boxes are often used to provide semantic information with minimal annotation effort.
However, in the context of endoscopic video, the elongated and articulated structure of surgical tools makes bounding boxes less informative.
They indeed often cover large portions of the image and significantly overlap with each other, reducing usefulness for precise localisation. 
We argue that skeletal pose annotations, such as those used in the field of human pose estimation\cite{cao2017realtime}, strike a better balance between semantic information and ease of annotation.
Furthermore, skeletal pose annotations offer the additional benefit of capturing 
structural information, since they can help localise the tip and the shaft area, as well as
instance-related information, since they can effectively distinguish between different instances of tools based on their unique skeletal structures.
Thereby, they provide richer and more precise insights compared to traditional bounding box annotations.

Peng et al.~\cite{peng2022autonomous} explore an alternative representation using tool tip bounding boxes combined with a line segment pointing along the tip.
Backer et al.~\cite{de2022multicentric} propose a method using vector annotation to create detailed wireframes for surgical instruments.
Instead of defining lines, different keypoints along the instruments are marked, allowing for precise representation of the instrument's structure and interactions within the surgical scene.
Du et al.~\cite{du2018articulated} provide tool pose annotations for 1,155 images from RMIT~\cite{10.1007/978-3-642-33418-4_70} and 1,850 from EndoVis~\cite{reiter2012feature}.
These two datasets are limited due to their small size and significant redundancy, due to being tightly sampled from the source videos.
In parallel to our efforts, Ghanekar et al.~\cite{ghanekar2025video} proposed a multi-frame, context-driven model for video-based tracking of surgical tool keypoints. 
Their approach segments keypoint regions across consecutive frames using optical flow and monocular depth as auxiliary cues, followed by centroid estimation for localisation. 
This method demonstrated accurate tracking performance across challenging datasets such as EndoVis 2015~\cite{du2018articulated} and JIGSAWS~\cite{gao2014jhu}, highlighting the benefits of temporal context in surgical tool-tip estimation.
Wu et al.~\cite{11127958} introduced SurgPose to support more generalisable pose estimation in robotic surgery. 
The dataset comprises over 120k annotated instances across six types of da Vinci instruments, each with seven semantic keypoints, collected using a novel UV-based labelling method. 
It also includes stereo image pairs, kinematic data, and joint states, enabling both 2D and 3D pose estimation.
SurgPose provides a strong foundation for vision-based pose estimation in robotic surgery, although it is currently limited to ex vivo environments and does not yet include complex scenes with mutual occlusions or inter-instrument interactions.
The PhaKIR (\url{https://phakir.re-mic.de})
% \href{https://phakir.re-mic.de}{The PhaKIR challenge}
, reported in ~\cite{rueckert2025comparativevalidationsurgicalphase}, advanced research in surgical instrument analysis by introducing a real-world multi-centre dataset with joint annotations for instance segmentation, keypoint estimation, and surgical phase recognition.
However, the keypoint estimation task proved particularly challenging, with only two teams submitting and both achieving limited performance, largely due to instrument variation, occlusion, and class imbalance.
This further motivated the design of ROBUST-MIPS to better support research on robust and generalisable tool pose estimation.

We aim to better establish the subfield of tool pose estimation by releasing ROBUST-MIPS (Medical Instrument Pose and Segmentation), a larger and more varied tool pose dataset, providing pose annotations for all 10,040 images of the ROBUST-MIS (Medical Instrument Segmentation) dataset~\cite{ross2021comparative, maier2021heidelberg}.
As each frame also has tool instance segmentations, we hope that this could be used to investigate the strengths and weaknesses of these two annotation approaches as well as the interplay between these two tasks.

\section*{Methods}
% This section describes the skeletal representation of various surgical instruments, organised into three components: data sources, annotation protocols, and technical validation.
We outline the methodology for creating the skeletal pose representation of various surgical instruments.
%according to three main components: data sources, labelling protocol, annotation software, and annotation procedures.
%
% The data sources subsection outlines the origin and application context of the raw data.
% The annotation protocols section details the number of keypoints defined for each instrument, along with their 2D coordinates and visibility annotations.
% Finally, the technical validation assesses the dataset's practical utility and its generalisability to skeletal pose estimation tasks.
%
First, we present the data sources, as well as the origin and application context of the raw images used for annotation. 
Next, we detail the labelling protocol, including the number and type of keypoints defined for each instrument, along with their types and visibility labels. 
We then discuss the software tool and techniques employed to address annotation challenges. 
Finally, we provide a comprehensive description of the annotation process, covering the step-by-step procedures and information about the annotators involved in creating the ROBUST-MIPS dataset.

\subsection*{Data Sources}
The ROBUST-MIPS dataset is derived from the ROBUST-MIS dataset, which was created for the ROBUST-MIS 2019 challenge~\cite{ross2021comparative}. 
This challenge aimed to benchmark algorithms for instrument segmentation and detection in minimally invasive surgery.
%(MIS). 
It comprises 10,040 laparoscopic frames extracted from 30 colorectal surgical procedures, including 10 rectal resections, 10 proctocolectomies, and 10 sigmoid resections, all performed at Heidelberg University Hospital. 
All data were acquired using a Karl Storz laparoscopic camera system and downsampled to 960x540 pixels for computational efficiency.
Ethical and legal considerations were addressed by fully anonymising all images, making the dataset suitable for public release without additional ethics approval~\cite{ross2021comparative}.

Frames were sampled at 1 frame per second, with additional frames extracted during surgical phrase transitions to ensure sufficient coverage of varying surgical contexts. 
Each frame was provided with a pixel-wise instance segmentation mask indicating surgical instruments.
ROBUST-MIS intentionally includes challenging imaging conditions typical of real-world surgical scenarios, such as bleeding, smoke, illumination changes, overlapping instruments, and partially visible tools.

\begin{figure}[bth]
\centering
\includegraphics[width=0.9\linewidth]{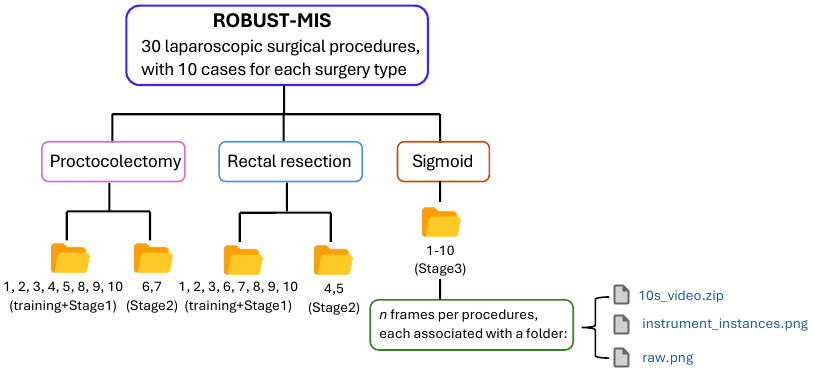}
% \caption{Legend (350 words max). Example legend text.}

\caption{\revmod{Overview of ROBUST-MIS data, the source data for the proposed ROBUST-MIPS. 
The directory structure is derived from the original ROBUST-MIS dataset~\cite{Roß_Reinke_Maier-Hein_Kopp-Schneider_Wagner_Kenngott_Müller-Stich_2019}. 
While the original dataset provides a 10-second video snippet (250 frames) with the last raw frame with its instance segmentation mask, the proposed ROBUST-MIPS dataset extends this structure by incorporating skeletal pose annotations. The final extended directory structure of our contribution is detailed in \figref{fig:MIPSstructure}.}\robustrevref{2}{1}}
\label{fig:fig1}
\end{figure}

The dataset was structured to support multiple tasks, including binary segmentation, multi-instance detection, and multi-instance segmentation, and was divided into training and testing sets to facilitate evaluation under increasing domain shifts.
The organisation of files and folders in the dataset is illustrated in \figref{fig:fig1}.
The dataset includes a detailed split strategy for accessing algorithm performance.
In particular, the testing set was split into three stages reflecting escalating domain gaps: Stage1 used data from the same patients as training, Stage2 from new patients but the same surgery type, and Stage3 from a different surgery type.

\subsection*{The Labelling Protocol}
To support the development of surgical instrument skeletal pose estimation, we have enriched the ROBUST-MIS dataset with additional annotations specifically designed for our target task.
To ensure consistent and generalisable annotations, we established a rigorous labelling protocol covering three main topics: keypoint selection, annotation scope, and annotation guidelines.

\subsubsection*{Keypoints Selection}
\begin{figure}[tbh]
\centering
\revref{2}{3a}
\revmod{\includegraphics[width=0.9\linewidth]{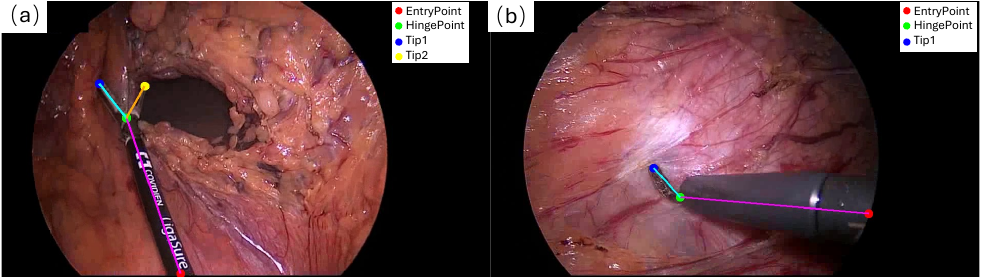}}
\caption{
Examples of selecting keypoints for different types of surgical instruments.
(a) Keypoints selected for an articulated surgical instrument.
(b) Keypoints selected for a rigid surgical instrument.}
\label{fig:fig2}
\end{figure}

In the skeletal pose representation method, keypoint selection is guided by the type and characteristics of surgical tools. In our ROBUST-MIPS dataset, the keypoints are categorised into 4 main types:
\begin{itemize}
\item \texttt{EntryPoint}: In minimally invasive surgery, images captured through an endoscope typically have a circular content area~\cite{budd2023rapid}. 
The intersection point between the surgical instrument shaft and the circular content area boundary is defined as the \texttt{EntryPoint}.
These points are represented by red dots in \figref{fig:fig2}.
\revref{2}{2a}
\revnew{Unlike structural landmarks (e.g., the \texttt{HingePoint}), the \texttt{EntryPoint} is not a fixed location on the instrument but varies dynamically as the tool moves in and out of the field of view (FoV).
This definition is consistent with previous work in surgical pose estimation~\cite{du2018articulated}.}
\item \texttt{HingePoint}: For rigid surgical instruments, the intersection between the shaft and the metal or plastic tip is defined as the \texttt{HingePoint}, as shown in \figref{fig:fig2}(b). For articulated surgical instruments, the joint is considered the \texttt{HingePoint}, as shown in \figref{fig:fig2}(a). These points are indicated by green dots in \figref{fig:fig2}.
\item \texttt{Tip1}/\texttt{Tip2}: 
The endpoints of all instruments can be labelled as tip points. 
For rigid instruments, there is only one endpoint, labelled as \texttt{Tip1}, as shown in \figref{fig:fig2}(b). 
For articulated instruments, there are two possible cases for endpoints: one endpoint, labelled as \texttt{Tip1}, or two endpoints, \revmod{arbitrarily} labelled as \texttt{Tip1} and \texttt{Tip2}.
\revref{2}{2b}
\revmod{Taking graspers as a typical example, due to the structural symmetry of the jaws and their continuous axial rotation during surgery, defining a fixed \emph{left} or \emph{right} orientation is ambiguous.
Even in cases where the two tips could be disambiguated (e.g. curved scissors), there is no universal convention across instruments to consistently order the tips.
Furthermore, distinguishing subtly differing tips becomes unreliable under occlusion, smoke or rapid motion. 
Enforcing a semantic tip distinction in such cases would introduce significant label noise without adding much value to the geometric understanding.
Consequently, \texttt{Tip1} and \texttt{Tip2} are defined as an unordered set in our dataset. 
Therefore, although \texttt{Tip1} and \texttt{Tip2} are distinguished by blue and yellow dots in \figref{fig:fig2} for visualization, this assignment is interchangeable and permutation-invariant.}
% \item \texttt{Tip1}/\texttt{Tip2}: 
% The endpoints of all instruments can be labelled as tip points. 
% For rigid instruments, there is only one endpoint, labelled as \texttt{Tip1}, as shown in \figref{fig:fig2}(b). 
% For articulated instruments, there are two possible cases for endpoints: one endpoint, labelled as \texttt{Tip1}, or two endpoints, labelled as \texttt{Tip1} and \texttt{Tip2}.  
% Since surgical tools do not have natural left and right elements, endpoints are represented by numbers, such as \texttt{Tip1} or \texttt{Tip2}, rather than being labelled as left or right as would be typical for human limbs. \texttt{Tip1} points are represented by blue dots, while \texttt{Tip2} points are represented by yellow dots in \figref{fig:fig2}.
\end{itemize}

\begin{figure}[!t]
\centering
\includegraphics[width=0.9\linewidth]{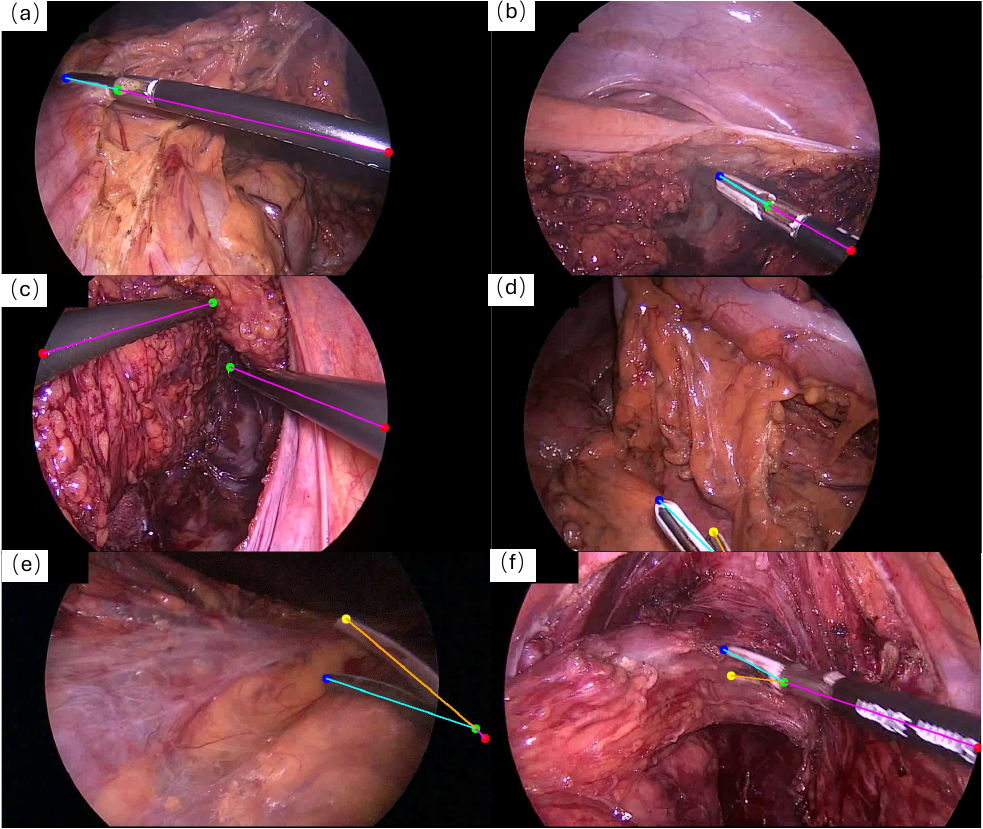}
\caption{
Examples of selecting valid keypoints in different visibility. 
(a) Selection of keypoints for partially occluded articulated surgical tools, \revnew{where one tip point is considered to be in an unpredictable missing state.} 
(b) Selection of keypoints for articulated surgical tool in a closed state, \revnew{where the two tips are considered to be at the same position. The second tip is labelled as \texttt{missing}.}
(c) Selection of keypoints for surgical tools with only the shaft visible in the FoV, \revnew{resulting in \texttt{missing} labels for both tips.} 
\revmod{(d) Selection of keypoints when the instrument shaft extends beyond the image boundary. 
To maintain skeletal connectivity, both the \texttt{HingePoint} and \texttt{EntryPoint} are annotated in the padding area with out-of-bounds coordinates, despite being strictly invisible.
(e) Selection of keypoints where the \texttt{HingePoint} is masked by the circular FoV but remains within the image frame. 
The point is geometrically predicted from the tips and arm structure, possessing valid positive coordinates.
In this case, the \texttt{HingePoint} and \texttt{EntryPoint} are both labelled as occluded.}
%\robustrevref{2}{2c}
(f) Selection of keypoints where one of the tips is inferred based on the instrument’s structural characteristics, such predicted keypoints are annotated as occluded.}
\marginnote{$R_{1}C_{1}$\\
$R_{2}C_{2c}$}
\label{fig:visiblity}
\end{figure}

\figref{fig:fig2} provides schematic examples of an articulated surgical tool and a rigid surgical tool, each illustrating the typical placement of these keypoints.
For articulated instruments such as bipolar clamps, blunt graspers, and scissors, which have tips that can open and consist of two parts (shaft and tip), four keypoints are typically annotated: \texttt{EntryPoint}, \texttt{HingePoint}, \texttt{Tip1}, and \texttt{Tip2}. 
Rigid instruments like dissection hooks and probes also consist of a shaft and a tip, but their tips cannot open, resulting in three annotated keypoints: \texttt{EntryPoint}, \texttt{HingePoint}, and \texttt{Tip1}. 
Importantly, in our ROBUST-MIPS dataset, every surgical instrument is annotated using the same four keypoint categories: \texttt{EntryPoint}- \texttt{HingePoint}-\texttt{Tip1}-\texttt{Tip2}.
However, the visibility status of each keypoint is crucial and is explicitly stored for each frame. 
\reversemarginpar
%\revref{1}{1}
\marginnote{$R_{1}C_{1}$\\
$R_{2}C_{2c}$}
\revmod{
The visibility labels are categorized into three states based on the visibility and inferability of the keypoint:
\begin{itemize}
    \item \texttt{visible}: The keypoint is clearly visible in the image.
    \item \texttt{occluded}: The keypoint is not directly visible (e.g., covered by tissue or located slightly outside the FoV) but its position can be reliably inferred based on the instrument's mostly rigid geometry or symmetry.
    \item \texttt{missing}: The keypoint is completely out of view with insufficient cues for inference, or physically does not exist (e.g., the second tip of a rigid tool).
\end{itemize}
While the definition of \texttt{visible} keypoints appears straightforward, the assignment of \texttt{missing} and \texttt{occluded} labels in specific scenarios can be practically challenging depending.
Illustrative cases are shown in \figref{fig:fig2} and \figref{fig:visiblity}, detailed as follows:

\begin{itemize}
    \item Scenarios for \texttt{missing}: 
    This label is applied when a keypoint is physically absent or cannot confidently be inferred by the annotator. In terms of geometric constraints, since rigid instruments lack a second tip (cf. \figref{fig:fig2} (b)) and closed articulated instruments have overlapping tips (cf. \figref{fig:visiblity} (b)), the fourth keypoint is marked as \texttt{missing} to maintain the data format. Regarding visual ambiguity, when the distal end is entirely hidden and visual cues are insufficient for inference by the annotator, the keypoints are defined as \texttt{missing}. This includes cases where the tip is entirely hidden (cf. \figref{fig:visiblity} (a)) or only the shaft is visible (cf. \figref{fig:visiblity} (c)).

    \item Scenarios for \texttt{occluded}: 
    This label applies to keypoints that are physically present but visually occluded. 
    For geometric inference within the image frame, if a tip is covered by tissue but inferable via symmetry (cf. \figref{fig:visiblity} (f)), or if a keypoint falls into the non-informative image region outside the circular FoV but remains within the rectangular image boundary (cf. \figref{fig:visiblity} (e)), they are labelled as \texttt{occluded} with valid positive coordinates.
    To maintain the connectivity of the annotation chain, the \texttt{EntryPoint} serves as the root node, it also be labelled as \texttt{occluded} in \figref{fig:visiblity} (e). 
    For skeletal connectivity beyond the image boundary, we allow the annotator to label points outside of the image boundary. As illustrated in \figref{fig:visiblity} (d), the instrument shaft may indeed be inferred to be completely outside the frame, meaning both the \texttt{EntryPoint} and \texttt{HingePoint} are located outside the image boundary.
    To accommodate this, our annotation software provides a zoom-out function and a designated padding area, allowing these points to be annotated with out-of-bounds coordinates.
    While these points ensure structural completeness, they are effectively filtered out during training due to their out-of-bounds values.
\end{itemize}}

\tabref{tab:representation} summarises the skeletal representations of surgical instruments, combining these variations in instrument types with differences in operational or visibility states.

\begin{table}[bth]
\centering
\caption{\label{tab:representation} Overview of the skeletal representation of articulated and rigid surgical tools in different states, with the visualisation of each case shown in \figref{fig:fig2} and \figref{fig:visiblity}.}
\revref{1}{2}
\begin{tabular}{|l|c|c|c|}
\hline
Tool types & States & Tool representation & Cases \\
\hline
\multirow{4}{*}{Articulated} & all keypoints  visible \revdel{(v-value > 0)} & \multirow{2}{*}{4 points and 3 lines} 
& \figref{fig:fig2}(a) \\
& \revnew{Tips or HingePoint occluded } & & \revnew{\figref{fig:visiblity}(d,e,f) }\\
                              & one tip missing/closed & 3 points and 2 lines & \figref{fig:visiblity}(a,b) \\
                              & only shaft in the FoV & 2 points and 1 line & \figref{fig:visiblity}(c) \\
\hline
\multirow{3}{*}{Rigid} & all keypoints visible \revdel{(v-value > 0)} 
& 3 points and 2 lines & \figref{fig:fig2}(b) \\
                       & only shaft in the FoV & 2 points and 1 line & similar with the \figref{fig:visiblity}(c) \\
\hline
\end{tabular}
\end{table}

\subsubsection*{Annotation Scope}
The annotation scope defines the entities to be annotated and excluded to ensure a focused and high-quality dataset suitable for surgical instrument pose estimation.
While the ROBUST-MIS dataset provides instance segmentation masks for various surgical tools and has been widely used for surgical scene analysis, adapting it for pose estimation revealed certain limitations in its original annotation protocol, particularly regarding the labelling of trocars and cannulas/ports.

In the context of segmentation, trocar cannulas are tubular devices that serve as ports during laparoscopic surgery and are considered visible structures that must be separately identified from surrounding tissues and surgical instruments.
Accordingly, the ROBUST-MIS dataset annotates both camera trocar cannulas, which hold the endoscope, and tool trocar cannulas, which serve as entry points for surgical instruments, as individual instances.

However, for pose estimation tasks, this level of detail introduces challenges:
\begin{itemize}
    \item Camera trocar cannulas: These are static structures fixed to the patient or surgical robot, contributing no dynamic motion or orientation information relevant for instrument pose estimation.
    \item Tool trocar cannulas: Although physically connected to the instruments, tool trocar cannulas merely represent a fixed entry point into the surgical field and do not reflect the dynamic geometry or movement of the instrument itself.
\end{itemize}

\begin{figure}[tbh]
\centering
\includegraphics[width=1.0\linewidth]{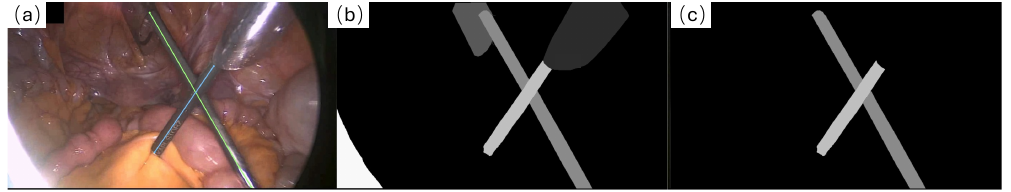}
\caption{
\revnew{Handling of tool trocar cannulas in ROBUST-MIPS annotations. 
(a) Pose annotation example in the presence of a tool trocar cannula. 
The distal end of the cannula is identified as the \texttt{EntryPoint}, while the instrument shaft extends to the \texttt{HingePoint}.
(b) The original segmentation annotation in ROBUST-MIS~\cite{Roß_Reinke_Maier-Hein_Kopp-Schneider_Wagner_Kenngott_Müller-Stich_2019}, where the trocar cannula is labelled as a distinct instance. 
(c) The refined segmentation annotation in ROBUST-MIPS dataset~\cite{Han_Budd_Zhang_Tian_Bergeles_Vercauteren_2025}, where the trocar cannula mask is removed to focus solely on the surgical instrument.
}
}
\marginnote{$R_{1}C_{3}$\\
$R_{2}C_{2d}$}
\label{fig:trocarmask}
\end{figure}

Including trocar cannulas in pose annotations can therefore introduce unnecessary noise and redundancy, as the functional pose of surgical instruments is defined by the shaft and tip beyond the trocar cannula rather than the trocar cannula itself.
When an instrument extends from a trocar cannula, we define the distal end of the trocar cannula as the \texttt{EntryPoint} for pose annotation, \revnew{as shown in \figref{fig:trocarmask} (a),} while other annotation principles remain unchanged.
For consistency, in the instance segmentation annotations of ROBUST-MIPS, we removed the masks corresponding to both camera trocar cannulas and tool trocar cannulas that were present in the original ROBUST-MIS dataset. \revnew{Taking the tool trocar cannulas as an example, \figref{fig:trocarmask} (b) and (c) illustrate this process by comparing the original instance label with our refined mask.
It should also be noted that frames containing only camera trocar cannulas do not possess corresponding skeletal annotations for surgical instruments.
Consequently, removing the camera trocar masks from these images results in completely empty (black) segmentation labels. }\revref{1}{3} \\\revref{2}{2d}
%Our custom annotation software enables efficient mask removal, ensuring that the resulting dataset is better suited for surgical instrument pose estimation tasks.

\subsubsection*{Annotation Guidelines}
ROBUST-MIS intentionally includes challenging imaging conditions typical of real-world surgical scenarios, such as bleeding, smoke, illumination changes, overlapping instruments, and partially visible tools. 
To address these challenging cases, we provided specialised instructions and examples for annotators. 
These included guidance on reviewing the corresponding 10-second video clips for clarification, referring to already annotated segmentation masks from ROBUST-MIS when uncertain, annotating as much of the instrument as possible, and continuing annotations even under low visibility conditions.
In some cases where it is not possible to distinguish between the shaft and the tip, or where no clear boundary exists between these parts, only the \texttt{EntryPoint} and \texttt{HingePoint} keypoints are annotated. 
Furthermore, additional factors such as motion blur, reflections, lens dirtiness, and the presence of fluids were also considered during annotation, ensuring robust and consistent labelling across diverse surgical scenes.

\subsection*{Annotation Software}
To support the annotation process for our ROBUST-MIPS dataset, we designed open-source annotation software specifically for manual surgical instrument pose labelling\revmod{, with its source code available on GitHub(\url{https://github.com/cai4cai/tool-pose-annotation-gui}).}
% \href{https://github.com/cai4cai/tool-pose-annotation-gui}{GitHub}.} 
The software provides a graphical interface that enables efficient image browsing and intuitive keypoint annotation.
\revnew{
Specifically, semantic abbreviations (e.g., `E' for \texttt{EntryPoint}, `T1' for \texttt{Tip1}) are displayed next to each point to aid in identifying instrument parts.}\revref{1}{4a}
Users can zoom out with the mouse scroll wheel. 
This function is particularly useful for placing occluded keypoints located outside the visible image area.
Annotation begins with a left click, which either starts a new pose or adds a keypoint to the current one. 
Right click is used to annotate occluded keypoints by placing an estimated position.
Middle click completes the current pose annotation. 
Clicking on the edge of an existing skeleton allows users to insert a visible/occluded transition point, and the remaining keypoint tags are automatically updated.
Our custom annotation software also enables efficient mask removal to ensuring that the instance segmentation masks are better suited for surgical instrument pose estimation tasks.
This software ensures efficient and consistent annotations across various surgical instrument types and visibility conditions, as described above.

\subsubsection*{Data Description}
\begin{figure}[tbh]
\centering
\includegraphics[width=0.9\linewidth]{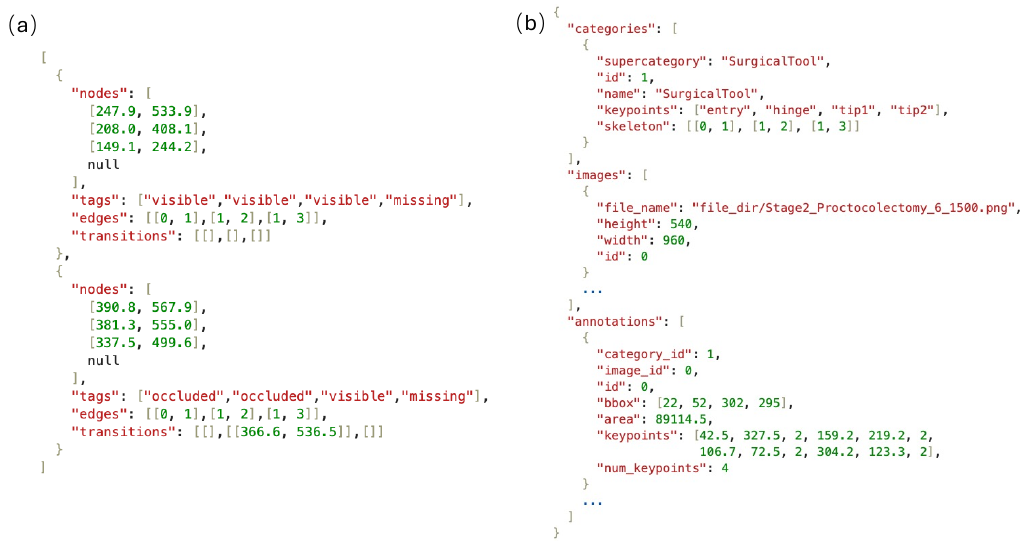}
\caption{(a) Example of JSON
annotation file from the custom annotation software. 
(b) Example of an annotation converted to the Microsoft COCO schema~\cite{lin2014microsoft} which allows for broad compatibility with human pose learning framework.}
\label{fig:jsonformat}
\end{figure}

The keypoint information obtained through the annotation software is stored as a JSON file using the schema shown in \figref{fig:jsonformat}, with one such JSON file generated for each image. 
In this schema, \texttt{nodes} contains the coordinates of the keypoints, while \texttt{tags} records their visibility status. The \texttt{edges} field specifies the connections between pairs of keypoints. 
The \texttt{transitions} field represents intermediate points located between visible keypoints and non-visible keypoints (either occluded or missing). 
As illustrated in \figref{fig:jsonformat}(a), if a tip point is visible but its connected hinge point is not, there must be a segment between them that is visible in the image. 
In this case, the farthest visible point along the arm from the tip is recorded as a \texttt{transition} point.

\subsection*{Annotation Procedures}
This section provides an overview of the workflow adopted to construct the ROBUST-MIPS dataset, along with insights into the human annotation process.
The description is organised into two main topics: The step-by-step procedure and the role of annotators.

\subsubsection*{Step-by-Step Procedure}
\begin{enumerate}
    \item Base dataset selection and preparation: 
    %(1 weeks): 
    The ROBUST-MIPS dataset was developed based on the existing open-access dataset ROBUST-MIS, which provided raw laparoscopic video data without instance-level keypoint annotations.
    \item Development of the labelling protocol:
    %(3 weeks): 
    Prior to annotation, we conducted extensive discussions to design a comprehensive labelling protocol. 
    This protocol specified definitions for keypoint selection and rules for handling challenging scenarios, such as partial occlusion, poor lighting conditions, and camera trocar cannulas.
    \item Manual annotation of keypoints:
    %(4 weeks): 
    All keypoint annotations were performed entirely manually using a custom annotation software. 
    Beyond spatial coordinates, the visibility status of each keypoint (visible, occluded, missing) was incorporated into the annotation process. 
    Annotators labelled each frame individually, specifying the positions of predefined keypoints for visible instruments. 
    In cases where certain keypoints were not visible, their status was recorded as either occluded or missing, following the defined protocol. 
    This additional information is crucial for downstream analysis and model training.
    \item Final quality control:
    %(2 weeks): 
    Upon completion of annotations, thorough quality control procedures were carried out.
    Each annotated frame was reviewed manually to verify correctness and consistency with the labelling protocol. 
    Additionally, the entire dataset underwent a second round of review by a different annotator to ensure accuracy and to resolve any potential discrepancies.
\end{enumerate}

\subsubsection*{Annotators}
The annotation process for creating the ROBUST-MIPS dataset involved a primary annotator responsible for the majority of annotation tasks and quality control. 
A secondary annotator, with greater experience,
%and medical qualifications, 
assisted in annotation and quality control.
Both annotators were supported by an expert team to ensure accuracy and consistency in the annotations.

\section*{Data Records}
\begin{figure}[tbh]
\centering
\includegraphics[width=0.8\linewidth]{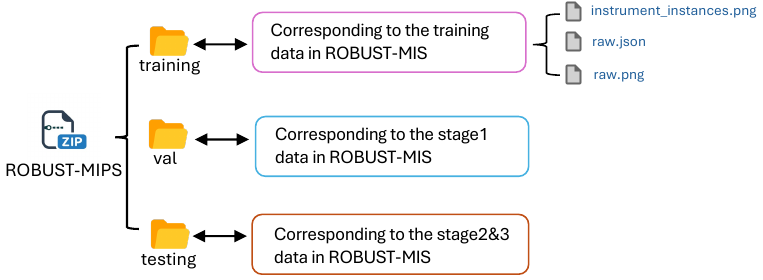}
\caption{Directory structure for ROBUST-MIPS dataset. Each split contains subdirectories following a \texttt{Surgery\_type/ Procedure\_ID/Frame\_ID} structure.
}
\label{fig:MIPSstructure}
\end{figure}

\revmod{The ROBUST-MIPS dataset~\cite{Han_Budd_Zhang_Tian_Bergeles_Vercauteren_2025} is accessible via the public repository on Synapse. The imaging data utilized in this work derive from the publicly available ROBUST-MIS dataset~\cite{Roß_Reinke_Maier-Hein_Kopp-Schneider_Wagner_Kenngott_Müller-Stich_2019}, which is also hosted on Synapse.}
% \revmod{The ROBUST-MIPS dataset~\cite{Han_Budd_Zhang_Tian_Bergeles_Vercauteren_2025} is accessible via the public repository on \href{https://doi.org/10.7303/syn64023381}{Synapse}. 
% The imaging data utilized in this work derive from the publicly available ROBUST-MIS dataset~\cite{Roß_Reinke_Maier-Hein_Kopp-Schneider_Wagner_Kenngott_Müller-Stich_2019}, which is also hosted on \href{https://doi.org/10.7303/syn18779624}{Synapse}.}
Our derived dataset is distributed as a compressed archive file, \texttt{ROBUST-MIPS.zip}.
The downloaded data requires a restructuring process to be reassigned according to the organization illustrated in \figref{fig:MIPSstructure}.
In this reorganized format, each frame consists of three components: the raw endoscopic image (\texttt{raw.png}), the corresponding instance segmentation mask (\texttt{instrument\_instances.png}), and the keypoint annotation file (\texttt{raw.json}).

To enhance the generalisability of the dataset and to facilitate subsequent performance evaluations on various popular pose estimation models, as discussed in the \emph{Models} Section, we processed the annotation information corresponding to each image and consolidated it into a JSON format similar to that used in the Microsoft COCO (Common objects in context) dataset~\cite{lin2014microsoft}, as illustrated in \figref{fig:jsonformat}(b).
In our dataset, all instruments are grouped into a single category, as shown in the \texttt{categories} object. 
The \texttt{images} object contains information about the file paths of each image, along with their resolutions and unique identifiers. 
The \texttt{annotations} object comprises a list in which each entry corresponds to the keypoint annotations for a specific frame. 
The \texttt{image\_id} corresponds to the id in the \texttt{images} object, while the \texttt{id} refers to the identifier of the current surgical tool.
Keypoint annotations are expressed as $(x, y, v)$, where $x$ and $y$ correspond to the horizontal and vertical coordinates of the keypoint, with the origin of the coordinate system at the top left of the image. 
$v$ indicates the visibility attribute of the keypoint. 
For each annotated keypoint, the value of the visibility property indicates if a keypoint is annotated and visible ($v = 2$), annotated and occluded ($v = 1$), or not annotated because it is not located inside the frame or in case it is not possible to estimate its position accurately ($v = 0$). 
The \texttt{num\_keypoints} represents the number of keypoints with a $v$-value greater than 0.

Additionally, the JSON file also includes the bounding box information for each instrument, which is calculated based on the coordinates of the keypoints and is denoted as $[x_{min}, y_{min}, w, h]$. 
The \texttt{area} object is computed as the square of the diagonal of the bounding box.

%%%%%
\subsubsection*{Bounding Box Generation}

\begin{figure}[tbh]
\centering
\includegraphics[width=0.7\linewidth]{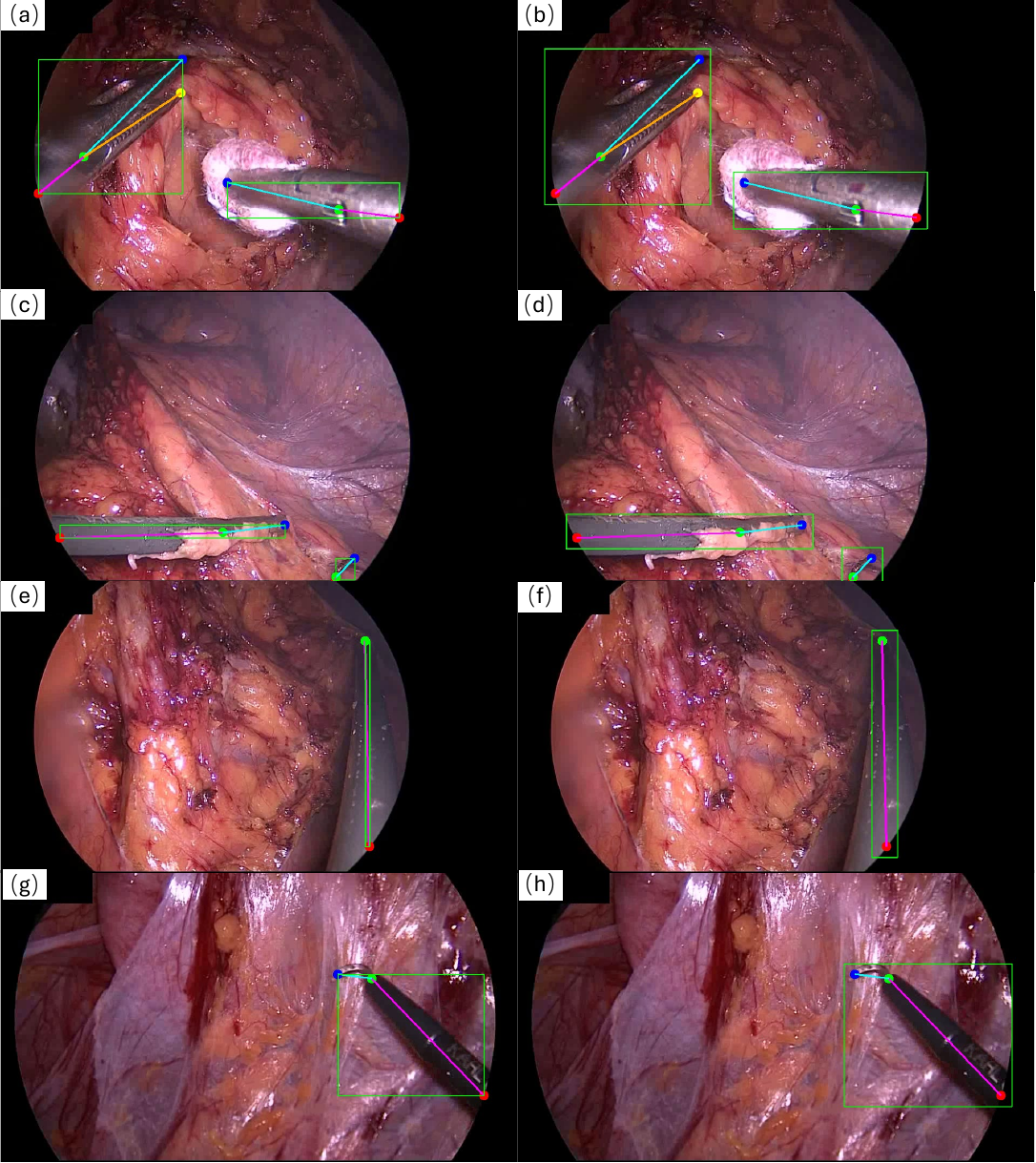}
\caption{Examples of bounding box generated from 2D keypoints. 
(a) Bounding boxes that can generally represent the surgical tool. 
(b) Result after adding a margin to the bounding box in (a). 
(c) Bounding box result for a surgical tool in a horizontal position within the FoV. 
(d) Result after adding a margin to the bounding box in (c). 
(e) Bounding box result for a surgical tool in a vertical position within the FoV. 
(f) Result after adding a margin to the bounding box in (e). 
(g) Bounding box result for a surgical tool with a curved shape. 
(h) Result after adding a margin to the bounding box in (g).}
\label{fig:bbox}
\end{figure}

To enable our dataset to support training based on both the top-down and bottom-up paradigms for pose estimation tasks~\cite{zheng2023deep}, the JSON files also include the bounding boxes calculated from the 2D keypoints for each surgical tool.
The top-left corner of each bounding box is defined by the minimum $x$ and $y$ coordinates, denoted as $(x_{min}, y_{min})$. 
The width $w$ and height $h$ of the bounding box are calculated as the differences between the maximum and minimum $x$ coordinates, $x_{max}-x_{min}$, and the maximum and minimum $y$ coordinates, $y_{max}-y_{min}$, respectively. 
Thus, the bounding box for each tool can be accurately represented as $[x_{min}, y_{min}, w, h]$. 
While the method of generating bounding boxes based on skeletal information effectively represents the pose of surgical instruments in most cases, as shown in \figref{fig:bbox}(a), it performs poorly when the surgical tool is in a horizontal or vertical position within the FoV, as illustrated in \figref{fig:bbox}(c,e). 
This is because the vertical or horizontal coordinates of the keypoints used for the calculation are too close to each other, resulting in overly narrow bounding boxes.
Additionally, for some tools with curved shapes, the bounding boxes calculated solely based on keypoint coordinates may not adequately represent the entire tool, as shown in \figref{fig:bbox}(g).

To address this issue and improve the accuracy of the bounding boxes generated from the 2D keypoint annotations, a margin of 20 pixels is added to the calculated boundaries on all sides. 
If the expanded bounding box exceeds the image boundaries, the image boundaries are used as the limit.

\subsubsection*{Dataset Split}
In the ROBUST-MIS Challenge 2019, the dataset was divided as shown in \tabref{tab:tab1} to evaluate the generalisability and performance of algorithms.
The ROBUST-MIPS dataset has similarly been structured but has been partitioned into training, validation, and testing sets, as shown in \tabref{tab:tab2}. 
It is important to note that images in Stage1 originate from the same patients as those in the training set; therefore, Stage1 data is used as the validation set in the ROBUST-MIPS dataset. 
Data from Stage2 and Stage3 have been allocated to the testing set to enable a comprehensive evaluation of the model generalisation ability.

\begin{table}[tbh]
\centering
\caption{\label{tab:tab1}Case distribution of the data with frames per stage and surgery of the ROBUST-MIS dataset. Empty frames (denoted as ef in the table) were classed as the \% of frames in which an instrument did not appear.}
\begin{tabular}{|l|c|c|c|c|}
\hline
\multirow{2}{*}{Procedure} & \multirow{2}{*}{Training} & \multicolumn{3}{c|}{Testing} \\ \cline{3-5} & & Stage 1 & Stage 2 & Stage 3 \\ 
\hline
Proctocolectomy & 2,943(2\% ef.) & 325 (11\% ef.) & 225 (11\% ef.) & 0 \\
\hline
Rectal resection & 3,040 (20\% ef.) & 338 (20\% ef.) & 289 (15\% ef.) & 0 \\
\hline
Sigmoid resection & 0 & 0 & 0 & 2880 (23\% ef.) \\
\hline
Total & 5983 (17\% ef.) & 663 (15\% ef.) & 514 (13\% ef.) & 2880 (23\% ef.) \\
\hline
\end{tabular}
\end{table}

\begin{table}[tbh]
\centering
\caption{\label{tab:tab2}Case distribution of the data with frames per stage and surgery of ROBUST-MIPS dataset. The training and validation data come from the same group of patients undergoing two types of surgeries, while the testing set includes data from different patients undergoing the same two surgery types, as well as a third surgery type not present in the training process.}
\begin{tabular}{|l|c|c|c|}
\hline
Procedure & Training & Validation & Testing \\
\hline
Proctocolectomy & 2,943 & 325 & 225 \\
\hline
Rectal resection & 3,040 & 338 & 289 \\
\hline
Sigmoid resection & 0 & 0 & 2880 \\
\hline
Total & 5983 & 663 & 3394 \\
\hline
\end{tabular}
\end{table}

\section*{Technical Validation}
This section presents a comprehensive overview and technical validation of our ROBUST-MIPS dataset, including showcases and performance evaluation. The performance evaluation results demonstrate its reliability and effectiveness for use in research and development of pose estimation models.

\subsection*{Training and Evaluation of Baseline Models}
\subsubsection*{Models}
\label{sec:PoseEstimationModels}
To validate the usability of the ROBUST-MIPS dataset, three baseline pose estimation models, RTMPose~\cite{jiang2023rtmpose}, SimpleBaseLine~\cite{xiao2018simple}, and ViTPose~\cite{xu2022vitpose} were trained.
To establish benchmark performance metrics for future researchers to use as a comparison, we chose to utilise a range of baselines
%or state-of-the-art (SOTA) models 
in pose estimation tasks.
Although these models were originally designed for human pose estimation, we employed them to explore their generalisability in surgical tool pose estimation.
Each keypoint of a surgical tool can be regarded as a joint in human pose estimation tasks. 
However, when annotating the human skeleton, symmetric points such as the left and right shoulders, or the left and right elbow joints, have distinct physical meanings, and consistency in annotation must be maintained across different individuals. 
In contrast, for symmetric surgical tools like scissors or forceps, the two tip points, \texttt{Tip1} and \texttt{Tip2}, do not require a specific order in the annotation.
When using the aforementioned models for surgical tool pose estimation, the equivalence between the tips of the surgical tool has not been addressed.

The models were trained using the training and validation splits of the ROBUST-MIPS dataset, with hyperparameters detailed in \tabref{tab:tab3}. 
All training was conducted on an NVIDIA A100 32G GPU. 
The MMPose open-source tool~\cite{mmpose2020} was utilised for both training and evaluation, as it includes a comprehensive set of relevant models and evaluation metrics.
For data augmentation, the MMPose framework implements standard techniques such as cropping, flipping, color distortion, rotation, and scaling.

\revref{2}{3b}
\revnew{Regarding the optimization objectives, we utilized the specific loss implementations provided by the MMPose framework. 
RTMPose was trained using KLDiscretLoss, which computes the KL divergence between predicted and ground-truth distributions on discretized coordinates \cite{li2022simcc}. 
In contrast, SimpleBaseLine and ViTPose utilized KeypointMSELoss. 
Unlike standard MSE or KL loss functions, these framework-specific implementations inherently incorporate a target weight masking mechanism. 
This ensures that keypoints annotated as \texttt{missing} are effectively excluded from backpropagation, thereby preventing invalid gradients from affecting the training process.}

\begin{table}[tbh]
\centering
\caption{\label{tab:tab3}Parameters of the models.}
\begin{tabular}{ll}
\toprule
optimiser           & AdamW \\
base learning rate  & 0.0005 \\
learning rate schedule & LinearLR \\
batch size             & 32(train) \\
                                & 16(val) \\
warm-up iterations      & 500 \\
weight decay           & 0.01 \\
training epochs        & 600 \\
\bottomrule
\end{tabular}
\end{table}

\subsubsection*{Recommended Metric}

The COCO Object Keypoint Similarity (COCO OKS) metric~\cite{lin2014microsoft} is designed to provide a quantitative assessment of the similarity between predicted keypoints and ground truth keypoints, taking into account the scale of the object and the relative importance of different keypoints:
\begin{equation}
OKS = \sum_{i}[\exp(-\frac{d_{i}^{2}}{2s^{2}\kappa_i^{2}})\delta(v_{i}>0)] / \sum_{i}[\delta(v_{i}>0)]
\label{equOKS}
\end{equation}
where $d_{i}$ is the Euclidean distance between the predicted keypoint and the ground truth keypoint. 
$\kappa_i$ is a per-keypoint constant that controls falloff, which helps in normalising the effect of different keypoints.
$v_{i}$ is the visibility flags of the ground truth (the predicted visibility tags are not used). Predicted keypoints that are not labelled ($v_{i}=0$) do not affect the OKS.
In \eqref{equOKS}, $s$ is the scale of the object.

As discussed above, the two tip points, \texttt{Tip1} and \texttt{Tip2}, do not require a specific order in the annotation.
In the COCO OKS, the equivalence between the tips of the surgical tool is not addressed.
We propose a simple modification of the metric where a version of the ground truth pose is constructed by swapping the order of the two tips.
We evaluate the OKS of a prediction against both the initial and tip-swapped ground truth and report the best value.
This achieves the same outcome as including a bipartite matching step as suggested in the PhaKIR challenge \cite{rueckert2025comparativevalidationsurgicalphase}.

\revref{1}{6}
\revmod{
In standard COCO OKS, $s$ is defined as the square root of the object bounding box area ($s=\sqrt{wh}$).
However, as shown in \figref{fig:bbox}, surgical tools are typically slender, elongated structures with high aspect ratios (length $\gg$ diameter).
For such objects, the area of an axis-aligned bounding box is highly sensitive to 2D rotation. It collapses to near-zero when the tool is axis-aligned (horizontal or vertical) but expands significantly when rotated diagonally.
Using the standard definition would thus result in an inconsistently strict metric, penalising axis-aligned poses disproportionately due to this area collapse.
To address this, we redefine $s$ based on the arithmetic mean of the squared dimensions (scaled diagonal):
\begin{equation}
s = \sqrt{\frac{w^2 + h^2}{2}}
\label{equ:s}
\end{equation}
Based on the inequality of arithmetic and geometric means, this formulation ensures $s^2 \ge wh$, providing a robust scale factor that is dominated by the tool's length (diagonal) rather than its projection width. This guarantees rotation invariance and prevents the evaluation scale from becoming excessively small in axis-aligned states.
For square-like objects (where $w \approx h$), the formula $\frac{w^2 + h^2}{2}$ mathematically reduces to the standard area ($w \cdot h$). This ensures that our metric remains consistent with the original definition of $s$ for non-slender shapes.}
% In COCO OKS, $s$ is typically defined as the square root of the object bounding box area ($s=\sqrt{wh}$).
% However, as shown in \figref{fig:bbox}, the area of the bounding box for surgical tools varies significantly across different operational states. 
% If we were to use the standard definition of $s$ from COCO OKS, it would introduce substantial errors when evaluating different images.
% To better evaluate the performance of the models in surgical tool pose estimation, we modified the COCO OKS metric to consider a more representative scale parameter.
% Specifically, we modified $s$ to be related to the diagonal of the bounding box:
% \begin{equation}
% s = \sqrt{\frac{w^2 + h^2}{2}}
% \label{equ:s}
% \end{equation}

\revref{1}{7}
\revmod{In the COCO human pose dataset, for each keypoint type $i$, a standard deviation $\sigma_i$ is defined to capture the expected human annotation uncertainty. 
These $\sigma_i$ are also used to score the quality of automatically predicted keypoints against manual annotations.
This value essentially reflects the difficulty of consistently labelling a specific anatomical landmark on the human body. 
For instance, distinct and well-localised landmarks like eyes are assigned a low scale-normalized $\sigma_i$ ($\approx 0.026$), meaning that predictions must be very precise relative to the object size to score highly. 
Conversely, structurally ambiguous points or those covered by soft tissue, such as hips, are assigned a higher $\sigma_i$ ($\approx 0.107$) which allows for a larger margin of error~\cite{lin2014microsoft}. 
To make COCO OKS a perceptually meaningful metric, the keypoint-specific falloff constants are typically set as $\kappa_i = 2\sigma_i$.

In our work, we lack the large-scale repeated annotations required to empirically calculate $\sigma_i$ for each specific instrument keypoint type. 
However, surgical tools present significant challenges, including diverse structural variations, tissue occlusion, and a lack of distinct surface textures (e.g., smooth metallic shafts), which inherently increase annotation variance. 
Therefore, we adopted a conservative strategy by relying on the maximum standard deviation $\sigma_i$ used in the COCO human pose dataset. 
We thus set $\sigma_i = 0.107$ (corresponding to the `hips' category) for all keypoints in our surgical tool annotation task. 
This choice acknowledges the geometric ambiguity of surgical tools and ensures that the evaluation metric remains fair and robust even under challenging visual conditions.}

\revmod{We utilize the official COCO implementation (\url{https://github.com/cocodataset/cocoapi/blob/master/PythonAPI/pycocotools/cocoeval.py}) for calculating the Average Precision (AP) and Average Recall (AR) metrics in keypoint detection.}
% \revmod{We utilize the official COCO implementation \href{https://github.com/cocodataset/cocoapi/blob/master/PythonAPI/pycocotools/cocoeval.py}{here} for calculating the Average Precision (AP) and Average Recall (AR) metrics in keypoint detection.}
AP measures the precision of the model in detecting keypoints, and AR evaluates the model ability to recall objects over a range of thresholds.
\textbf{AP$_{k}$} represents the average precision at a given OKS threshold $k$, while \textbf{AR$_{k}$} denotes the average recall at OKS threshold $k$.
The default \textbf{AP(OKS)} and \textbf{AR(OKS)} are generally averaged over multiple OKS values, specifically from 0.50 to 0.95 with increments of 0.05.

Additionally, since Top-Down methods rely on a two-stage pose estimation model, the first stage involves bounding box detection. 
The evaluation of this stage aligns with the evaluation methods used in object detection tasks, utilising the COCO Intersection over Union (IoU) metric. 
IoU measures the overlap between the predicted bounding box and the ground truth bounding box, calculated as the ratio of the area of intersection to the area of union. 
Likewise, IoU plays a key role in calculating AP and AR. 
\textbf{AP$_{\text{IoU}=k}$} represents the average precision at a given IoU threshold $k$, and \textbf{AP(IoU)} is the COCO-style metric, averaged over IoU thresholds from 0.5 to 0.95 in increments of 0.05.
\textbf{AR$_{\text{IoU}=k}$} denotes the average recall at IoU threshold $k$, and \textbf{AR (IoU)} follows the same COCO-style averaging process.

\subsubsection*{Performance Evaluation}
\begin{figure}[!tbp]
\centering
\includegraphics[width=0.9\linewidth]{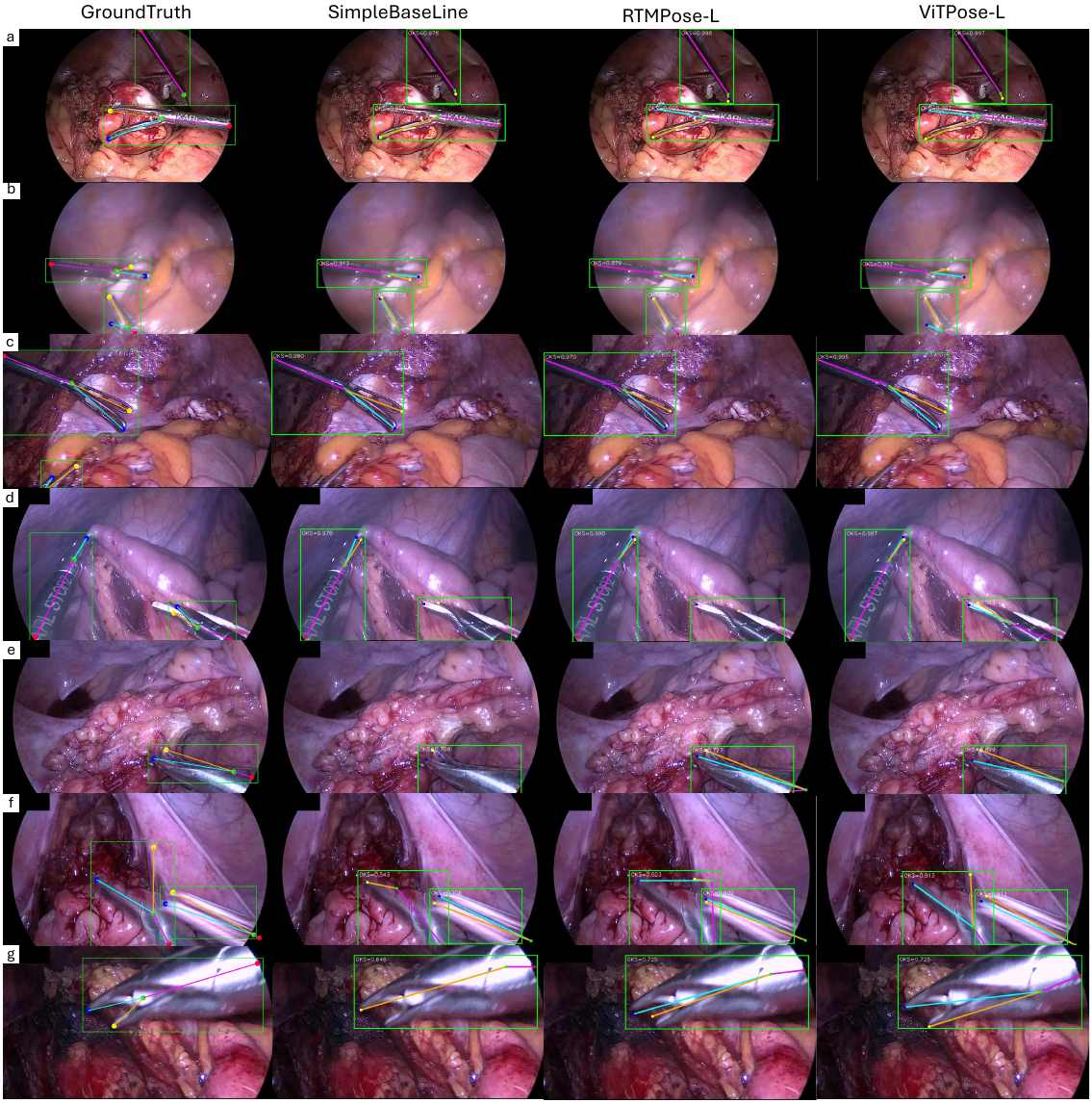}
\caption{\revmod{Visual representation of the performance of SimpleBaseLine, RTMPose, ViTPose and the corresponding ground truth annotations. OKS score values are indicated on each subfigure and these are better visualised by digital zoom in. Predicted keypoints that are labelled as missing ($v_{i}=0$) do not affect the OKS.}\robustrevref{2}{4}}
\label{fig:fig3}
\end{figure}

The performance results are presented in \tabref{tab:tab4}, demonstrating the effectiveness of the dataset in training robust pose estimation models.
The models were evaluated on the testing set that was not previously encountered during training or validation to assess their generalisation capability.  
A qualitative comparison between the model predictions on the testing set and the corresponding ground truth can be seen in \figref{fig:fig3}.
% A video of more results from those models on the testing dataset can be found at \url{http://todo}.

\begin{table}[tbhp]
\centering
\caption{\label{tab:tab4} Results of various algorithms for surgical tool pose estimation on the ROBUST-MIPS testing set. SBL stands for SimpleBaseLine~\cite{xiao2018simple}.}
\begin{threeparttable}
\begin{tabularx}{\textwidth}{|c|c|c|X|X|X|X|X|X|}
\hline
\multirow{2}{*}{Model}& \multirow{2}{*}{Backbone} & \multirow{2}{*}{Resolution} & \multicolumn{6}{c|}{Robust-MIP testing} \\ \cline{4-9} 
& & & \textbf{AP} & \textbf{AP$_{\text{OKS}=0.5}$} & \textbf{AP$_{\text{OKS}=0.75}$} & \textbf{AR} & \textbf{AR$_{\text{OKS}=0.5}$} & \textbf{AR$_{\text{OKS}=0.75}$} \\ \hline
SBL & ResNet152 & 256x192 & 0.694 & 0.819 & 0.704 & 0.732 & 0.834 & 0.739 \\ \hline
SBL & ResNet152 & 384x288 & 0.684 & 0.807 & 0.694 & 0.730 & 0.830 & 0.740 \\ \hline
RTMPose & CSPNext-m & 256x192 & 0.705 & 0.820 & 0.716 & 0.740 & 0.839 & 0.748 \\ \hline
RTMPose & CSPNext-l & 256x192 & 0.712 & 0.827 & 0.722 & 0.750 & 0.845 & 0.758 \\ \hline
% HRNet & HRNet-W32 & 256x192 & 0.622 & 0.794 & 0.623 & 0.693 & 0.816 & 0.695 \\ \hline
% HRNet & HRNet-W32 & 384x288 & 0.626 & 0.792 & 0.626 & 0.695 & 0.815 & 0.699 \\ \hline
% UDP & HRNet-W32 & 256x192 & 0.627 & 0.79 & 0.63 & 0.698 & 0.816 & 0.702 \\ \hline
% UDP & HRNet-w32 & 384x288 & 0.626 & 0.792 & 0.626 & 0.695 & 0.815 & 0.699 \\ \hline
% HRFormer & HRFormer-B & 256x192 & 0.632 & 0.803 & 0.629 & 0.702 & 0.82 & 0.705 \\ \hline
ViTPose-B & ViT-B & 256x192 & 0.735 & 0.832 & 0.750 & 0.768 & 0.847 & 0.778 \\ \hline
ViTPose-L & ViT-L & 256x192 & \textbf{0.754} & \textbf{0.842} & \textbf{0.771} & \textbf{0.784} & \textbf{0.855} & \textbf{0.796} \\ \hline
\end{tabularx} 
%\begin{tablenotes}
%\item[1] SBL represents SimpleBaseLine.
%\end{tablenotes}
\end{threeparttable}
\end{table}

\subsection*{Limitations and Possible Improvements}
The proposed dataset in this paper has a few limitations that must be taken into consideration.
One of the main limitations is that not all surgical tools can be accurately represented using this scheme. 
For example, curved instruments like scissors or hooks present challenges. 
In the case of scissors, the shaft is not straight, so the line segment connecting the keypoints does not accurately represent the actual shape of the tool. 
Similarly, for hooks, the line connecting the tip and the hinge point fails to capture the curvature of the hook.
Another limitations is that all the surgical tools are categorised under a single class without more detailed classification labels, such as forceps, hooks, scissors, needle drivers and so on.
While this may be sufficient for the task of surgical instrument pose estimation, it limits the generalisability of the dataset.

In addition, the baseline models employed in this study predict the endpoints of surgical instruments independently, and the annotation process does not enforce a consistent order of these points across samples.
As a result, the predicted endpoints may occasionally be assigned in a different order from the reference annotation.
While the invariance with respect to tip order is already considered in our proposed modified OKS, the phenomenon nonetheless also underscores a modelling limitation.
Future architectures may benefit from explicitly encoding tip-level equivalence or ordering invariance during training, which could enhance both prediction stability and semantic consistency.

\section*{Usage Notes}
ROBUST-MIPS is released under a Creative Commons Attribution-NonCommercial ShareAlike license (CC BY-NC-SA)\revnew{, which is required to align with the non-commercial restrictions applied to the source data.}
\revdel{Any use or mention of this dataset must include a citation to this paper and the ROBUST-MIS~\cite{ross2021comparative, maier2021heidelberg}. 
If the dataset is utilised in the creation of new works, they should also include citations to this paper.
This licensing choice aligns with the release license of the ROBUST-MIS dataset, from which our annotations are derived, ensuring that the original license is retained and respected.}

% \section*{Data Availability}

% The ROBUST-MIPS dataset generated and analyzed in this study is publicly available at Synapse (\url{https://doi.org/10.7303/syn64023381}) and mirrored on \href{https://drive.google.com/file/d/1RiizCpgMx8OKo5n1GuiuzonqA4heKdqK/view?usp=share_link}{Google Drive}. 
% The imaging data used to construct this dataset were obtained from the publicly available ROBUST-MIS dataset, accessible via Synapse (\url{https://doi.org/10.7303/syn18779624}).

\section*{Data Availability}

The ROBUST-MIPS dataset generated and analyzed in this study is publicly available at \href{https://doi.org/10.7303/syn64023381}{https://doi.org/10.7303/syn64023381} ~\cite{Han_Budd_Zhang_Tian_Bergeles_Vercauteren_2025}. \revdel{and mirrored on \href{https://drive.google.com/file/d/1RiizCpgMx8OKo5n1GuiuzonqA4heKdqK/view?usp=share_link}{Google Drive}.} 
The imaging data used to construct this dataset were obtained from the publicly available ROBUST-MIS dataset, accessible via \href{https://doi.org/10.7303/syn18779624}{https://doi.org/10.7303/syn18779624} ~\cite{Roß_Reinke_Maier-Hein_Kopp-Schneider_Wagner_Kenngott_Müller-Stich_2019}.

\section*{Code Availability}
The annotation software is made public at \href{https://github.com/cai4cai/tool-pose-annotation-gui}{https://github.com/cai4cai/tool-pose-annotation-gui}. 
We also release the code for benchmark training at \href{https://github.com/cai4cai/ROBUST_MIPS_toolpose}{https://github.com/cai4cai/ROBUST\_MIPS\_toolpose}. 
It also contains scripts for converting the data to the COCO format.

\bibliography{sample}

% \noindent LaTeX formats citations and references automatically using the bibliography records in your .bib file, which you can edit via the project menu. Use the cite command for an inline citation, e.g. \cite{Kaufman2020, Figueredo:2009dg, Babichev2002, behringer2014manipulating}. For data citations of datasets uploaded to e.g. \emph{figshare}, please use the \verb|howpublished| option in the bib entry to specify the platform and the link, as in the \verb|Hao:gidmaps:2014| example in the sample bibliography file. For journal articles, DOIs should be included for works in press that do not yet have volume or page numbers. For other journal articles, DOIs should be included uniformly for all articles or not at all. We recommend that you encode all DOIs in your bibtex database as full URLs, e.g. https://doi.org/10.1007/s12110-009-9068-2.

\section*{Acknowledgements}
\revnew{Data Sources: We would like to thank the authors of the ROBUST-MIS dataset for making their data publicly available, which served as the foundation for this work.
Funding Sources: }This work was supported by core funding from Wellcome/EPSRC [WT203148/Z/16/Z; NS/A000049/1].
Additional support was received from the European Union’s Horizon 2020 research and innovation programme under grant agreement No. 101016985 (FAROS project), and from Wellcome [WT223880/Z/21/Z].
% This work was supported by core funding from the Wellcome/EPSRC [WT203148/Z/16/Z; NS/A000049/1].
% %
% This project has received funding from the European Union's Horizon 2020 research and innovation programme under grant agreement No 101016985 (FAROS project).
%
For the purpose of open access, the authors have applied a CC BY public copyright licence to any Author Accepted Manuscript version arising from this submission.

\section*{Author Contributions Statement}
 
Zhe Han: Data curation, Methodology, Validation, Writing- Original draft preparation.
Charlie Budd: Software, Data curation, Writing- Reviewing and Editing.
Gongyu Zhang: Writing- Reviewing and Editing.
Huanyu Tian: Data curation, Writing- Reviewing and Editing.
Christos Bergeles: Supervision.
Tom Vercauteren: Conceptualisation, Supervision.

\section*{Competing Interests}
T.V. is a co-founder and shareholder of Hypervision Surgical Ltd, London, UK. The authors declare that they have no other
conflict of interest.

\end{document}